\documentclass{article}
\usepackage[margin=1in]{geometry}
\usepackage{times}
\usepackage{amsmath,amssymb,amsthm}
\usepackage{graphicx}
\usepackage{booktabs}
\usepackage{multirow}
\usepackage{xcolor}
\usepackage[hidelinks]{hyperref}
\usepackage{url}
\usepackage{natbib}
\usepackage{caption}
\usepackage{float}
\newcommand{\pp}{\,\text{pp}}
\newcommand{\TV}{\operatorname{TV}}

\newcommand{\qwen}{Qwen2.5}

\title{A Shared Learning Rate Is Not a Neutral Control\\
in Selective On-Policy Distillation}
\author{Chencheng Zhu\\
UNSW Sydney\\
\texttt{chencheng.zhu@student.unsw.edu.au}}
\date{}

\begin{document}
\maketitle

\begin{abstract}
Selective on-policy distillation trains a student only at the token positions
a selector scores highest, and the literature compares selectors under a
single shared learning rate---a control chosen to be neutral. We show it is
not. Under LoRA adaptation, across a $8\times$ learning-rate grid on GSM8K
with a \qwen{}-1.5B student and 7B teacher, dense supervision is
statistically flat (swing $1.8\pp$, $p{=}0.26$) while every selective arm we
test moves with the rate:
$5.4\pp$ for a \emph{random} $5\%$ subset, $6.7\pp$ for a total-variation
selector, $11.7$--$17.7\pp$ for a teachability selector. The asymmetry has a
direct consequence for how these methods are compared: the
dense-versus-selective verdict reads $10.1\pp$ at $\eta{=}10^{-4}$ and
$5.1\pp$ at $5{\cdot}10^{-5}$---the same comparison, differing by
$2.0\times$, decided by a parameter the protocol treats as scenery. Among
the selectors themselves, rankings stay stable in our setting but two of six
pairwise significance calls flip between adjacent rates---the protocol
changes what a paper concludes without any rank inversion. We call
this \emph{selector--rate entanglement} and trace it to selection itself
rather than to step size: AdamW update magnitudes track the rate to within
$2.2\%$ despite $15.5\times$ gradient-norm differences across arms. A
frozen-scoring ablation---selection scored by the initial student, with
criterion, budget, and on-policy rollouts unchanged---isolates how much of
the entanglement comes from selection reading the model it is training: in a
preregistered test at 12 seeds per cell, live scoring adds $3.79 \pm
1.69\pp$ of rate sensitivity ($p{=}0.035$) and produces a strictly separated
selection-drift trajectory, while the frozen arm remains significantly
entangled itself ($p{=}0.015$)---the loop aggravates the phenomenon rather
than causing it. The added sensitivity comes from the cool end of the grid,
where live scoring is $2.98\pp$ \emph{better} ($p{=}0.001$) rather than
worse, so the very same ablation reads as harmful or beneficial depending on
which single rate an experimenter fixes. Under full fine-tuning at the rates
this literature actually uses ($10^{-6}$--$10^{-5}$), the pattern survives in
graded form and grows: dense itself swings $19.8\pp$, the selective arm
$49.5\pp$ ($2.5\times$ more), and the dense-versus-selective verdict ranges
from a non-significant $+3.6\pp$ at $2{\cdot}10^{-6}$---the published
operating point---to $+34\pp$ ($p{=}0.005$) one notch hotter. On MATH-500
the rate dependence does not reproduce under LoRA, scoping that result,
while the cost of selective training there does (${\sim}10\pp$). We prescribe reporting the arm~$\times$~rate
matrix, not a shared-rate column, as a precondition for selector
comparisons.
\end{abstract}

%======================================================================
\section{Introduction}
\label{sec:intro}

\begin{figure}[t]
\centering
\includegraphics[width=.8\linewidth]{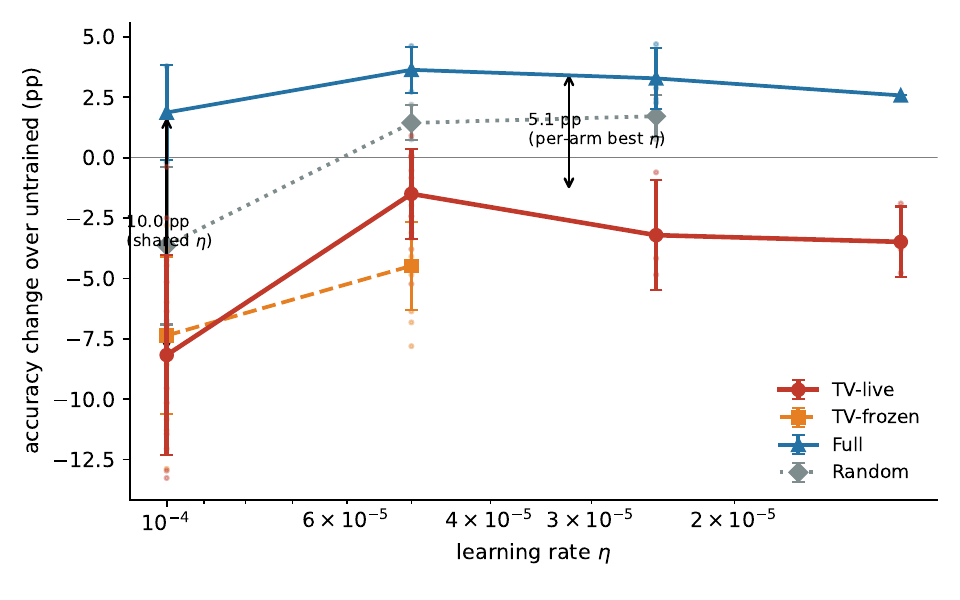}
\caption{\textbf{Dense supervision is flat across the learning-rate grid;
every selective arm is not.} Accuracy change over the untrained student
(GSM8K, mean $\pm$ seed SD; dots are seeds) versus learning rate. Swings:
dense $1.8\pp$ ($t{=}1.40$, n.s.), random $5\%$ subset $5.4$,
frozen-scored TV $2.9$ ($p{=}0.015$), live-scored TV $6.7$ ($p{<}0.001$).
The consequence for comparisons: the dense-versus-selective verdict is
$10.1\pp$ at $\eta{=}10^{-4}$ and $5.1\pp$ at $5{\cdot}10^{-5}$---a factor
of $2.0$ decided by a parameter the protocol treats as neutral.}
\label{fig:headline}
\end{figure}

To compare token selectors, the selective-distillation literature holds
everything else fixed: same data, same teacher, same token budget, and one
shared learning rate
\citep{wang2026teachability,xu2026tip,jiang2026rock,huang2025selectkd,
koo2025switch}. The protocol embodies an assumption so natural it is never
stated: that a selector is a static filter---an exogenous lens held up to
the data---so that whatever separates two arms trained under identical
hyperparameters is the selectors' contribution.

The assumption has a testable consequence: if the learning rate is neutral,
then arms should respond to it alike, and the ranking between them should be
stable across the grid. Figure~\ref{fig:headline} shows that they do not and
it is not. Under LoRA adaptation, across an $8\times$ grid, dense
supervision is statistically flat (swing $1.8\pp$, $t{=}1.40$) while every
selective arm moves: a
\emph{random} $5\%$ subset swings $5.4\pp$, a total-variation selector
$6.7\pp$, a teachability selector $11.7$--$17.7\pp$. Restricting the loss
support---even to a subset chosen without looking at anything---makes the
outcome a function of the learning rate in a way dense training is not. We
call this asymmetry \emph{selector--rate entanglement}. It is not a LoRA
artifact: under full fine-tuning at the rates this literature actually
publishes with, everything becomes rate-sensitive and the asymmetry survives
in graded form---the selective arm swings $2.5\times$ more than dense, and
the gap between them moves from statistically invisible at the published
rates to $34\pp$ one notch hotter (Section~\ref{sec:ft}).

The consequence for the comparison protocol is direct. The
dense-versus-selective verdict reads $10.1\pp$ at $\eta{=}10^{-4}$ and
$5.1\pp$ at $5{\cdot}10^{-5}$: the same two arms, the same data, the same
budget, a factor of $2.0$ between the two answers, and nothing to choose
between the rates except convention. A comparison run at one shared rate
does not report a property of the selector; it reports a property of the
selector \emph{at that rate}, and the literature does not currently vary the
rate (Table~\ref{tab:littable}).

An obvious mundane explanation must die first. Selective losses have very
different gradient scales, so perhaps the arms simply take different
effective steps at the same nominal rate. They do not: gradient norms differ
by $15.5\times$ across arms, yet the AdamW update magnitude per unit rate is
constant to within $2.2\%$ (Section~\ref{sec:notstepsize}). The arms take
equal-sized steps in different directions; no rescaling of the rate can
repair the comparison.

Where does the asymmetry come from? Selection in these methods is recomputed
from the \emph{current} student at every rollout refresh---a selection
frozen at initialization would be a different method---so the selector reads
the variable it is helping to move. Section~\ref{sec:openloop} tests whether
that dependence matters by freezing the scoring model at $\theta_0$ while
holding criterion, budget, and on-policy rollouts fixed. It changes the
\emph{dynamics of selection} unambiguously: frozen-scored selections drift
monotonically away from the initial selection while live-scored ones turn
around and re-converge, with no overlap across seeds. On accuracy, a
preregistered test at 12 seeds per cell puts the interaction at $3.79 \pm
1.69\pp$ ($p{=}0.035$): live scoring adds about four points of rate
sensitivity on top of what restricting the loss support already costs. The
frozen arm stays entangled itself ($2.89\pp$, $p{=}0.015$), so the scoring
loop aggravates the phenomenon without being its origin---and the added
sensitivity sits at the cool end of the grid, where live scoring is
$2.98\pp$ better rather than worse. We arrive at this
number the long way: an earlier version of this work argued the same point
from ``significant in one arm, not the other,'' which is not a valid
inference \citep{gelman2006difference}, and the correction required
quadrupling the sample (Appendix~\ref{app:prereg}, P7).

This paper contributes a measurement, an exclusion, and a protocol change.
We measure selector--rate entanglement on a duration-homogeneous
arm~$\times$~rate matrix spanning four selection rules, two datasets, two
adaptation regimes, and 133 training runs; we exclude the step-size explanation by logging actual
parameter displacement; and we show what the entanglement does to the
comparisons this literature reports, prescribing the arm~$\times$~rate
matrix in place of a shared-rate column (Section~\ref{sec:implications}).
Every decision criterion was fixed before its data, including the three that
went against us (Appendix~\ref{app:prereg}). The summary fits in one
sentence: a selector comparison that fixes the learning rate does not
control for it.

%======================================================================
\section{Setup: Selection as a Model-Dependent Component}
\label{sec:setup}

\subsection{Setup}
\label{sec:setup-details}

We study on-policy distillation of \qwen{}-1.5B-Instruct from
\qwen{}-7B-Instruct on GSM8K \citep{cobbe2021gsm8k,qwen2024qwen25}. Training
proceeds in \emph{refresh rounds}: in each round the student generates 512
fresh rollouts from training prompts (768 max new tokens, temperature $0.7$,
top-$p$ $0.9$), the teacher scores every generated position, a selector keeps
the top $5\%$ of positions, and the student takes 512 optimizer steps on the
distillation loss restricted to the selected positions (LoRA rank 32, AdamW,
gradient clipping at $1.0$; three rounds, 1{,}536 steps total). Selection is
recomputed at the start of every round from live student and teacher
scores---the granularity every selective-distillation method in this
literature shares, since a selection frozen at initialization would be a
different method from the one proposed. We evaluate greedily on all 1{,}319
GSM8K test problems and report accuracy change over the untrained student
($64.97\%$); greedy evaluation removes sampling noise from an already
noise-limited contrast. Full training details are in
Appendix~\ref{app:details}.

\subsection{The loop}
\label{sec:the-loop}

The selector we make closed-loop scores each position $i$ by the total
variation distance between the student's and teacher's next-token
distributions,
\begin{equation}
s_t(i) \;=\; \TV\!\big(p_{\theta_t}(\cdot \mid x_i),\;
q(\cdot \mid x_i)\big),
\qquad
M_t \;=\; \operatorname{top-5\%}(s_t),
\label{eq:score}
\end{equation}
and the update rule trains only on $M_t$:
$\theta_{t+1} = \mathcal{A}\big(\theta_t, \nabla \mathcal{L}|_{M_t}\big)$.
The composition is the point of this paper: $\theta$ determines $s$, $s$
determines $M$, and $M$ determines the next $\theta$. The selector is a
thermostat that also heats the room---it measures the very variable it is
helping to move. In control terms the training system is a closed loop, and
the learning rate $\eta$ is its gain: it sets how far the controlled variable
moves between two re-measurements.

Two arms in our design never read $\theta$ and therefore run open-loop:
\textsc{Full} trains on every position ($M_t = $ all), and \textsc{Random}
draws $M_t$ uniformly at the same $5\%$ budget. Both change other factors
relative to the TV arm as well (budget and criterion respectively);
Section~\ref{sec:openloop} constructs the comparison that changes nothing
else.

\subsection{Measuring the loop: drift observables}
\label{sec:drift-obs}

If the loop matters, its action should be visible on the selection itself,
not only on end-task accuracy. At every round we score the round's rollout
positions twice---once with the current model $\theta_t$ and once with the
frozen initial model $\theta_0$---and compare the two selections they induce:
the Spearman correlation $\rho_t$ of the two score vectors, and the Jaccard
overlap $J_t$ of the two top-$5\%$ sets. Because rollouts are regenerated
each round, positions have no identity across rounds; $J_t$ asks, \emph{on
today's data, do $\theta_t$ and $\theta_0$ still agree about what to select?}

$J_t$ is the primary observable. The top-$5\%$ boundary is thin: after a
single round of training, $\rho$ falls only $2.6\%$ (to $0.974$) while $J$
falls $43$--$48\%$ (to $0.52$--$0.57$), because small score perturbations
flip many positions near the threshold. Rank correlation is too blunt an
instrument for a $5\%$ tail; set churn is not. These within-run observables
average over ${\sim}512$ chains per round, giving them an effective sample
size that accuracy contrasts at $n{\leq}5$ seeds cannot approach.

%======================================================================
\section{Selector--Rate Entanglement}
\label{sec:entangle}

Table~\ref{tab:matrix} is the arm~$\times$~rate matrix underlying
Figure~\ref{fig:headline}. Every cell is duration-homogeneous (1{,}536
optimizer steps, selection recomputed once per round); an earlier version of
this matrix mixed training durations across columns and was discarded by
audit (Appendix~\ref{app:prereg}).

\begin{table}[t]
\centering
\caption{\textbf{Arm $\times$ learning-rate matrix (duration-homogeneous).}
Accuracy change (pp) over the untrained baseline ($64.97\%$, greedy, 1{,}319
problems), mean $\pm$ seed SD; parenthesized counts are seeds. Swing is the
difference between the arm's best and worst cell means. \textsc{TV-frozen} is
the frozen-scoring arm of Section~\ref{sec:openloop}.}
\label{tab:matrix}
\begin{tabular}{llrrrrr}
\toprule
Arm & Loop & $\eta{=}10^{-4}$ & $5{\cdot}10^{-5}$ & $2.5{\cdot}10^{-5}$ & $1.25{\cdot}10^{-5}$ & Swing \\
\midrule
\textsc{Full}      & --     & $+1.87{\pm}2.0$ (3) & $+3.64{\pm}1.0$ (3) & $+3.29{\pm}1.3$ (3) & $+2.58$ (1) & $1.8$ \\
\textsc{Random}    & open   & $-3.64{\pm}3.3$ (3) & $+1.44{\pm}0.7$ (3) & $+1.72{\pm}0.9$ (3) & -- & $5.4$ \\
\textsc{TV-frozen}   & $\theta_0$ & $-7.37{\pm}3.2$ (12) & $-4.48{\pm}1.8$ (12) & -- & -- & $2.9$ \\
\textsc{TV-live} & $\theta_t$ & $-8.18{\pm}4.1$ (12) & $-1.50{\pm}1.9$ (12) & $-3.21{\pm}2.3$ (3) & $-3.49{\pm}1.5$ (3) & $6.7$ \\
\bottomrule
\end{tabular}
\end{table}

\paragraph{Selective training is rate-entangled even without feedback---our
own strong hypothesis died here.}
We preregistered \textsc{Random} as a falsification test with frozen
thresholds: if the fully state-independent random selection swings more than
$4\pp$ across the grid, then rate sensitivity is not exclusive to the
feedback loop, and our initial hypothesis---that the loop is the sole
source---is falsified. It fired. \textsc{Random} swings $5.4\pp$ (every cell
$n{=}3$), while dense supervision is statistically flat ($1.8\pp$,
$t{=}1.40$, $p{=}0.26$). The criterion was written on swing magnitude; we
report the corresponding endpoint test as well ($+5.36 \pm 1.95$,
$t{=}2.74$, $p{=}0.097$)---above the preregistered threshold, below
conventional significance at three seeds. Merely restricting the loss to a
$5\%$ subset---\emph{any} subset---couples the outcome to the learning rate.
We report this as a first-class finding: the entanglement has a
loop-independent layer, and any account (including our original one) that
attributes it entirely to feedback is wrong.

\paragraph{Coupling depth orders the damage at the shared rate.}
The four arms differ in how much the selection reads the training state:
\textsc{Full} not at all; \textsc{Random} restricts support but ignores
state; \textsc{TV-frozen} applies a model-based criterion frozen at
$\theta_0$; \textsc{TV-live} applies the same criterion live. At
$\eta{=}10^{-4}$ the cell means fall in exactly this order---$+1.87 \to
-3.64 \to -7.37 \to -8.18\pp$---but only the endpoint contrast is
established (\textsc{Full} vs.\ \textsc{TV-live}: $10.05\pp$, $t{=}6.09$,
$p{<}0.001$). The individual rungs are not: $+5.51$ ($p{=}0.08$), $+3.73$
($p{=}0.17$), and, once the TV arms reach $n{=}12$, $+0.81$ ($p{=}0.60$).
The last rung in particular is indistinguishable from zero, so at the hot
rate the frozen and live scoring arms are equally damaged. We report the
ordering as a description of the means and rest the claim on the endpoints.
Section~\ref{sec:openloop} shows where the two TV arms \emph{do} separate:
at the cooler rate, not this one.

\paragraph{The verdict a shared-rate comparison returns depends on which
rate is shared.}
At $\eta{=}10^{-4}$ dense supervision beats the live-scored TV selector by
$10.05\pp$ ($t{=}6.09$); at $5{\cdot}10^{-5}$ the same contrast is
$5.14\pp$. Both are answers to ``how much does this selective recipe cost?''
computed on identical data with identical budgets; they differ by
$1.96\times$ (seed-level bootstrap $95\%$ CI $[1.27, 2.76]$, 20k resamples).
Neither rate is privileged---both lie inside the grid, and the published
LoRA range brackets them (Table~\ref{tab:littable}).

We state precisely what this is and is not. It is \emph{not} an estimate of
how much per-arm tuning would shrink the gap: our per-cell means come from
the same test set used to report accuracy, with no held-out split for
selecting a rate, and in fact both arms happen to peak at the same cell
($5{\cdot}10^{-5}$), so the two numbers above are two shared-rate columns
rather than a tuned-versus-untuned contrast. Quantifying a tuning protocol
properly requires a validation split and a declared tuning budget, which we
did not run. What the pair of numbers does show is narrower and still
consequential: the answer a single-shared-rate comparison returns is a
function of the rate chosen, by a factor of $2.0$ here. In a literature
where selector-versus-selector margins are routinely ${\sim}1$--$3\pp$, that
dependence is large enough to reorder methods, which is why we prescribe
reporting the matrix rather than a column.

\paragraph{What we retracted along the way.}
Two earlier versions of this section's claims did not survive our own
audits, and we record them because both failure modes are generic. At
$n{=}1$ we observed an apparent \emph{ranking inversion} between arms at
their best cells; it vanished at $n{=}3$. And our first matrix produced a
headline ratio of $5.2\times$ that was an artifact of mixed training
durations across columns---the shared-rate column came from $8\times$
shorter runs than the rest. Duration-homogeneous, the ratio is $2.0\times$.
A third retraction, of a causal claim about the scoring path, is recorded in
Appendix~\ref{app:prereg} (P7). Comparisons in this area are fragile enough
that the paper diagnosing them made three errors of the kind it diagnoses.

%======================================================================
\section{What It Is Not: Step Size}
\label{sec:notstepsize}

Before attributing the entanglement to a feedback loop, we must dispose of a
far more mundane candidate. A loss restricted to $5\%$ of positions has a
different gradient scale than a dense loss; if arms take different effective
step sizes at the same nominal $\eta$, then the ``entanglement'' would be
nothing but mis-scaled steps, and the fix would be a per-arm rescaling of
$\eta$ by gradient norm.

The premise is real and the conclusion is false, and both halves are
measurements. Pre-clipping gradient norms do differ dramatically across
arms---by a factor of $15.5$ ($0.35$ to $5.42$)---so the concern is not a
straw man. But the quantity that reaches the weights is not the gradient; it
is the AdamW update, and we log the actual per-step parameter displacement
for every run. The ratio of update norm to $\eta$ is constant across arms to
within $2.2\%$: AdamW's per-coordinate normalization erases the
gradient-scale differences before they touch a single weight. All arms take
the same size steps. They differ in \emph{where} the steps point---which is
determined by what the selector reads.

This measurement matters beyond hygiene: it eliminates the entire family of
``rescale the learning rate by the gradient norm'' remedies, which would
otherwise be the obvious response to our results. No rescaling of step size
can fix a confound that is not in the step size.\footnote{This was the first
of five mechanism hypotheses we tested; four died. (i) ``$12\times$ larger
gradients take $12\times$ larger steps''---killed by the $2.2\%$ dispersion
above. (ii) ``dense supervision wins because the global objective is
unbiased''---\textsc{Random} ranks mid-field, explaining only part of the
gap. (iii) ``discarding $95\%$ of rollouts is the main cost''---killed by a
direct budget-matched comparison. (iv) ``high-gradient-norm arms are hurt
more at large $\eta$''---killed by the same update/$\eta$ constancy. (v) The
drift-magnitude version of the loop hypothesis is killed in
Section~\ref{sec:openloop}; what survives is the drift-dynamics version.}

%======================================================================
\section{Does the Model-Dependence of Selection Matter?}
\label{sec:openloop}

\subsection{The intervention: freeze the scoring model}
\label{sec:intervention}

One cannot freeze the selection \emph{set}: rollouts are regenerated every
round, so positions have no identity across rounds and there is no set to
carry forward. What can be frozen is the \emph{scoring model}. Our
frozen-scored arm, \textsc{TV-frozen}, computes the same score as
Eq.~\ref{eq:score} but with $\theta_0$ in place of $\theta_t$;
\textsc{TV-live} is the method the literature describes. Rollouts remain
on-policy in both arms, the teacher is fixed in both, and criterion and
budget are identical. Implementation is free with LoRA: disabling the
adapter recovers $\theta_0$ exactly.

\paragraph{What this ablation removes, and what it leaves.} It cuts the
\emph{direct} path by which training rewrites its own selection scores. It
does not remove the model-dependence of selection entirely: rollouts are
still generated by $\theta_t$, so which positions are available to be
selected keeps tracking the current model, in both arms. We name the arm
\textsc{TV-frozen} rather than \textsc{TV-frozen} for exactly this reason, and
scope every claim below to the direct scoring path. Among the comparisons
available to us it remains the tightest: \textsc{Full} versus \textsc{TV}
changes budget \emph{and} scoring dependence; \textsc{Random} versus
\textsc{TV} changes criterion \emph{and} scoring dependence;
\textsc{TV-frozen} versus \textsc{TV-live} changes the scoring dependence
alone. We run both arms at $\eta \in \{10^{-4}, 5{\cdot}10^{-5}\}$, the two
rates spanning the live arm's swing.

\subsection{Preregistration}
\label{sec:prereg}

All decision criteria were frozen as executable assertions before the data
arrived (Appendix~\ref{app:prereg} gives them verbatim). Drift was declared
the primary observable; the accuracy contrast was declared underpowered in
advance (seed SDs of $2$--$4\pp$ at $n{\leq}5$) and demoted to directional
support. One deviation is disclosed: after the $n{=}3$ closed-vs-open
contrast at $\eta{=}10^{-4}$ landed at $t{=}{-}2.77$ against a threshold of
$2.78$, we extended the two closed cells---symmetrically, under rules frozen
before the new data: same Welch test, both sample sizes reported, no sampling
beyond $n{=}5$ regardless of outcome. Section~\ref{sec:signflip} reports what
the extension did to the estimate, which is itself instructive.

\paragraph{Live scoring adds rate entanglement, by $3.79 \pm 1.69\pp$.}
The quantity at issue is an interaction---does the arm's rate effect depend
on which model scores the selection---and we report it directly, because the
tempting shortcut of comparing one arm's significance to the other's is
invalid \citep{gelman2006difference}. At $n{=}12$ per cell
(Table~\ref{tab:s1}) the live-scored arm's rate effect is
$\Delta_{\text{live}} = 6.68 \pm 1.31\pp$ ($p{<}0.001$), the frozen-scored
arm's is $\Delta_{\text{frozen}} = 2.89 \pm 1.07\pp$ ($p{=}0.015$), and
their difference---the interaction---is $\mathbf{3.79 \pm 1.69\pp}$
($t{=}2.24$, $p{=}0.035$, $95\%$ CI $[+0.29, +7.30]$). Reading the loop
into selection scores therefore costs roughly $3.8\pp$ of additional rate
sensitivity beyond what restricting the loss support already costs.

\paragraph{The interaction is carried by the cool rate, not the hot one.}
Decomposing it by rate is instructive and, we think, the most interesting
thing in this section. At $\eta{=}10^{-4}$ the two arms are
indistinguishable ($-0.81 \pm 1.52\pp$, $p{=}0.60$): when the rate is hot
enough to hurt, it hurts both equally. At $5{\cdot}10^{-5}$ live scoring is
\emph{better} by $2.98 \pm 0.75\pp$ ($p{=}0.001$). Letting selection track
the model is therefore not a liability that a well-chosen rate mitigates;
it is a feature that only pays off at a well-chosen rate, and the same
intervention would be written up as harmful, neutral, or beneficial
depending on which single rate an experimenter happened to fix---this
paper's own thesis, applied to its own ablation. (These per-rate contrasts
are descriptive: the confirmatory test we preregistered is the interaction,
and an earlier per-rate analysis was capped at $n{=}5$ by its own stopping
rule, Appendix~\ref{app:prereg} P3.)

Three qualifications belong with the interaction estimate. First, the
confidence interval only just excludes zero; the effect is established, not
pinned down. Second, the frozen arm is itself significantly
rate-entangled ($p{=}0.015$), so the scoring path is an aggravating factor
and not the origin of the phenomenon---consistent with \textsc{Random}
in Section~\ref{sec:entangle}. Third, this test exists because an earlier
version of this paper made the invalid inference: at our original $n{=}5/3$
the interaction was $3.60 \pm 2.84$ ($p{=}0.25$), which establishes
nothing, and we argued the claim from significance-in-one-arm instead. The
preregistered replication (Appendix~\ref{app:prereg}, P7) fixed the sample
size at $n{=}12$ in advance with no top-ups. What makes the confirmation
credible is not the $p$-value but the stability of the point estimate as
$n$ grew: $3.60 \to 3.79$, with the interval shrinking around it.

\begin{figure}[t]
\centering
\includegraphics[width=.72\linewidth]{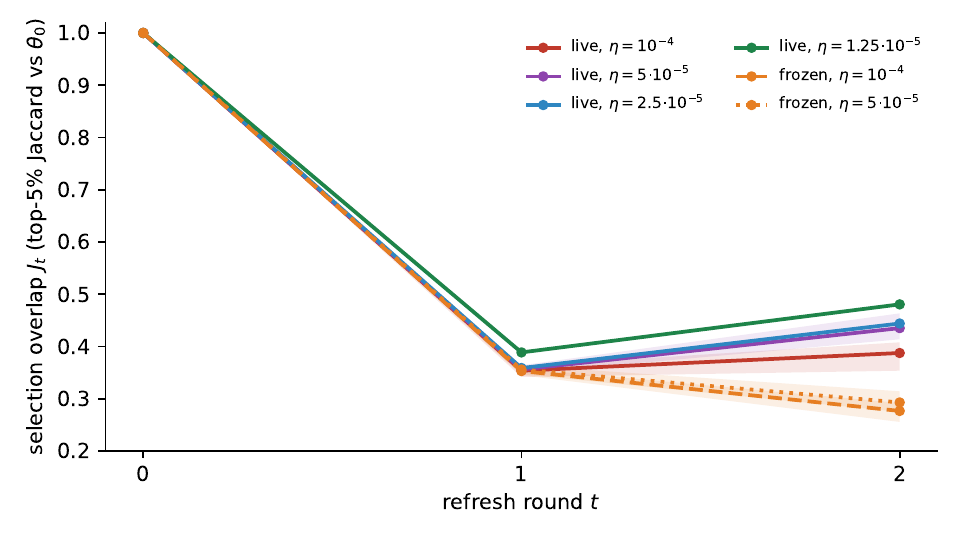}
\caption{\textbf{The loop changes the shape of selection drift, not its
magnitude---with a dose--response in the rate.} Top-5\% Jaccard $J_t$
between $\theta_t$- and $\theta_0$-scored selections on the same positions
(mean over seeds; bands are seed ranges; $n{=}12$ per TV cell at the final
round). After one round every arm sits at $J_1 \approx 0.35$ regardless of
rate or scoring source. Then they separate: frozen-scored selections keep
drifting away ($\to 0.28$--$0.29$, both rates), live-scored ones recover
toward the initial selection, with the recovery monotone in the rate
($J_2 = 0.39$, $0.44$, $0.44$, $0.48$ from $\eta{=}10^{-4}$ down to
$1.25{\cdot}10^{-5}$). The frozen and live ranges do not overlap at any
seed ($\max 0.314$ vs.\ $\min 0.354$; $t{=}17.6$).}
\label{fig:drift}
\end{figure}

\paragraph{The loop changes the shape of selection drift, not its magnitude.}
Figure~\ref{fig:drift} shows the $J_t$ trajectories, and they falsify our own
first mechanism hypothesis. The naive causal chain---larger $\eta$ moves
$\theta$ farther, drift grows, instability follows---predicts that drift
\emph{magnitude} tracks $\eta$. It does not: after one round, $J_1 \approx
0.35$ in every arm at every rate. What separates the arms is the second
round. Open-loop selections keep drifting away from $\theta_0$ ($J$:
$1.00 \to 0.35 \to 0.27$--$0.29$, both rates, indistinguishable). Closed-loop
selections turn around: $J$ recovers, and the recovery strength is monotone
in the rate across all four grid points ($J_2 = 0.39$, $0.44$, $0.44$,
$0.48$ from $\eta{=}10^{-4}$ down to $1.25{\cdot}10^{-5}$; the two
lowest-rate points are single runs). At $n{=}12$ the two scoring sources do
not overlap on a single seed ($t{=}17.6$, $p{<}10^{-13}$), which makes this
the most sharply separated measurement in the paper---and a reminder that a
within-run observable can be decisive where an end-task contrast at the same
seed count is not. The loop does not amplify drift; it
couples drift
\emph{dynamics}---recovery versus continued divergence---to the learning
rate. This is the observable on which the confound acts, and it is invisible
to any protocol that logs only end-task accuracy.

\paragraph{A directional observation: the sign of the loop's net effect
flips with the rate.}
\label{sec:signflip}
At $\eta{=}10^{-4}$ the loop hurts (closed$-$open $= -2.40 \pm 2.58\pp$,
$n{=}5$; $-5.43 \pm 1.96$ at $n{=}3$); at $5{\cdot}10^{-5}$ it helps
($+1.21 \pm 1.19\pp$). The signs are opposite at both sample sizes, but
neither end is individually significant, and by our preregistered rule this
is recorded permanently as directional support, not a finding. The $n{=}3
\to n{=}5$ shrinkage at $10^{-4}$ ($-5.43 \to -2.40$) is worth pausing on:
it is regression to the mean, caught because the extension rule was frozen
before the data and forced symmetric sampling. Had we extended only the
near-threshold cell, or stopped on significance, this paper would be
reporting an inflated effect---the same failure mode, at the meta level,
that it diagnoses in selector comparisons.

\begin{table}[t]
\centering
\caption{\textbf{Scoring selection with the live model adds $3.79\pp$ of
rate sensitivity.} Accuracy change (pp) over the untrained baseline, mean
$\pm$ seed SD. Both arms: TV criterion, $5\%$ budget, on-policy rollouts,
512 steps/round, selection recomputed every round; the arms differ only in
which model scores the selection. $\Delta(\eta)$ is the within-arm
difference across rates (Welch). The $n{=}12$ block is the preregistered
powered test (P7); the smaller-$n$ block is the original, underpowered
version, reported as preregistered.}
\label{tab:s1}
\begin{tabular}{lcrrr}
\toprule
Arm & Scoring & $\eta{=}10^{-4}$ & $\eta{=}5{\cdot}10^{-5}$ & $\Delta(\eta)$ \\
\midrule
\multicolumn{5}{l}{\emph{Preregistered powered test, $n{=}12$ per cell}}\\
\textsc{TV-frozen} & $\theta_0$ & $-7.37$ & $-4.48$ & $2.89 \pm 1.07$ ($p{=}0.015$) \\
\textsc{TV-live}   & $\theta_t$ & $-8.18$ & $-1.50$ & $6.68 \pm 1.31$ ($p{<}0.001$) \\
\multicolumn{4}{r}{\textbf{interaction}} & $\mathbf{3.79 \pm 1.69}$ ($p{=}0.035$) \\
\midrule
\multicolumn{5}{l}{\emph{Original underpowered version}}\\
\textsc{TV-frozen} ($n{=}3$) & $\theta_0$ & $-5.99 \pm 2.84$ & $-3.36 \pm 1.08$ & $2.63$ ($p{=}0.25$) \\
\textsc{TV-live} ($n{=}5$)   & $\theta_t$ & $-8.39 \pm 4.46$ & $-2.15 \pm 2.25$ & $6.23$ ($p{=}0.032$) \\
\multicolumn{4}{r}{interaction} & $3.60 \pm 2.84$ ($p{=}0.25$) \\
\bottomrule
\end{tabular}
\end{table}

\subsection{Replication with other criteria}
\label{sec:s2}

Does the loop effect generalize beyond the TV criterion? We preregistered
the sharpest version of the question (Appendix~\ref{app:prereg}): a
student-only \emph{entropy} selector also reads $\theta_t$, so if the loop
hypothesis holds as stated, it must entangle more when closed than open. We
ran the full open/closed design for Entropy and a closed-only arm for the
teachability criterion \citep{wang2026teachability}
($2$ rates $\times$ $3$ seeds each; Table~\ref{tab:s2}).

\begin{table}[t]
\centering
\caption{\textbf{S2 replication.} Accuracy change (pp), mean $\pm$ seed SD,
$n{=}3$ per cell. $J_1$ is the first-round selection drift (smaller = the
criterion's scores move more as the model moves).}
\label{tab:s2}
\begin{tabular}{llrrrr}
\toprule
Arm & Scoring & $\eta{=}10^{-4}$ & $5{\cdot}10^{-5}$ & $\Delta(\eta)$ & $J_1$ \\
\midrule
\textsc{Ent-live}   & $\theta_t$ & $+0.38{\pm}1.2$  & $+3.54{\pm}0.1$ & $3.16$ ($p{=}0.043$) & $0.62$ \\
\textsc{Ent-frozen}     & $\theta_0$ & $-0.48{\pm}2.0$  & $+3.82{\pm}0.3$ & $4.30$ ($p{=}0.059$) & $0.62$ \\
\textsc{Teach-live} & $\theta_t$ & $-15.54{\pm}3.2$ & $-3.87{\pm}1.1$ & $11.68$ ($p{=}0.016$) & $0.86$ \\
\textsc{Teach-frozen}   & $\theta_0$ & $-22.11{\pm}7.1$ & $-4.37{\pm}4.0$ & $17.74$ ($p{=}0.030$) & --- \\
\bottomrule
\end{tabular}
\end{table}

\paragraph{The preregistered verdict fired against us.}
Entropy entangles ($\Delta(\eta) = 3.16$, $p{=}0.043$)---but identically
with the loop open ($4.30$): the interaction is absent ($-1.14 \pm 1.34$),
and the drift-recovery signature does not reproduce (closed $J_2 > J_1$ in
only $1/3$ seeds at either rate, versus $5/5$ and $3/3$ cells for TV). By
our frozen criteria this is recorded as \textsc{revise-loop}: \emph{whatever
the frozen-scoring ablation of Section~\ref{sec:openloop} is detecting, it is
specific to the TV criterion and does not generalise to model-dependent
selection in general}. Meanwhile the teachability
selector swings $11.68\pp$ ($t{=}6.04$)---the largest of any arm we
measured, with a catastrophic $-15.5\pp$ at the shared rate.

\paragraph{A coupling account, stated post hoc and then tested once.}
The account was formed after the Entropy verdict: the criteria differ in
how much selection \emph{churn} training induces, measured by first-round
drift ($J_1 = 0.35$, $0.62$, $0.86$ for TV, Entropy, Teach), and churn is
the carrier of the loop effect---so the open/closed interaction should
shrink along that ordering. This makes a prediction for the one cell we had
not yet run: Teach, with the most stable selection, should show an
interaction indistinguishable from zero and below TV's $+3.60$. We froze
that criterion in a hash-stamped script and then ran \textsc{Teach-frozen}.
Measured interaction: $-6.07 \pm 5.06$ ($t{=}{-}1.20$; consistent with
zero, below $3.60$)---the account survives its first out-of-sample test,
with the caveat that the interval is wide. Strikingly, \textsc{Teach-frozen}
itself swings $17.74\pp$, reaching $-22.1\pp$ at the shared rate: the
criterion's rate fragility needs no loop at all. Two further observations
stand regardless of the account: drift magnitude does not predict damage
(Teach has the most stable selection and the largest swings), and every
selective criterion we tested is rate-entangled while dense supervision is
not.

\subsection{Replication on a second dataset: the damage transfers, the
entanglement does not}
\label{sec:m500}

Both results so far come from GSM8K. We repeated the decisive contrast on
MATH \citep{hendrycks2021math}---the numeric-answer subset, so the answer
checker and the accuracy definition are reused byte-identically and the
measuring instrument is not a variable---training on MATH prompts and
evaluating on the 325 numeric-answer problems of MATH-500
\citep{lightman2024verify}. Grid: $\{\textsc{Full}, \textsc{TV-live}\}
\times \{10^{-4}, 5{\cdot}10^{-5}\} \times 3$ seeds, everything else
unchanged; the untrained student scores $49.23\%$ on this set. The verdict
rule was fixed before the runs: entanglement counts as reproduced only if
\textsc{TV-live} shows a significant rate effect that also exceeds
\textsc{Full}'s.

\begin{table}[t]
\centering
\caption{\textbf{MATH-500 replication.} Greedy accuracy (\%) on the 325
numeric-answer problems, mean $\pm$ seed SD, $n{=}3$; the untrained student
scores $49.23\%$ on the same set, and the bracketed values are changes over
it.}
\label{tab:m500}
\begin{tabular}{lrrr}
\toprule
Arm & $\eta{=}10^{-4}$ & $5{\cdot}10^{-5}$ & $\Delta(\eta)$ \\
\midrule
\textsc{Full}      & $48.00 {\pm} 3.49$ \;($-1.23$) & $50.56 {\pm} 0.77$ \;($+1.33$) & $2.56$ ($t{=}1.24$, $p{=}0.33$) \\
\textsc{TV-live} & $37.95 {\pm} 2.69$ \;($-11.28$) & $39.69 {\pm} 2.77$ \;($-9.54$) & $1.74$ ($t{=}0.78$, $p{=}0.48$) \\
\midrule
gap & $10.05 {\pm} 2.55$ ($p{=}0.019$) & $10.87 {\pm} 1.66$ ($p{=}0.015$) & \\
\bottomrule
\end{tabular}
\end{table}

\paragraph{Verdict: not reproduced---the rate entanglement is scoped to
GSM8K.} On MATH, \textsc{TV-live}'s rate effect is $1.74\pp$
($p{=}0.48$) and does not exceed \textsc{Full}'s $2.56\pp$ ($p{=}0.33$);
neither arm is rate-sensitive over this range. By the frozen rule this is
recorded as \textsc{not reproduced}, and every entanglement claim in this
paper is scoped to the GSM8K setting accordingly. We flag one caveat that
cuts against over-reading the negative: with $325$ evaluation problems and
seed SDs of $2.7$--$3.5\pp$, this grid could not have detected a $2$--$3\pp$
rate effect, so it bounds the effect rather than excluding it.

\paragraph{What did transfer, and it is the more useful half.} The
dense-vs-selective damage replicates almost exactly: $10.05$ and $10.87\pp$
at the two rates on MATH, against $10.05\pp$ at the shared rate on GSM8K,
all significant. Measured against each dataset's own untrained student the
two settings are near-isomorphic: dense supervision is neutral-to-slightly-
positive ($-1.2$ to $+1.3\pp$ on MATH, $+1.9$ to $+3.6$ on GSM8K) while the
$5\%$ TV-selected budget costs about ten points ($-11.3$ to $-9.5$ on MATH,
$-8.2$ to $-1.5$ on GSM8K). Combined with the flatness of \textsc{Full} on
both datasets, this says the protocol
prescription of Section~\ref{sec:implications} is not a GSM8K artifact even
though the entanglement magnitude is: whichever dataset one works on, the
selective arm is the fragile one and the dense arm is the stable reference,
so the two must not be compared at a single shared rate without checking.

\subsection{Replication under full fine-tuning, at the literature's own
rates}
\label{sec:ft}

Everything so far is LoRA at rates $10$--$100\times$ hotter than the
full-fine-tuning rates the literature publishes with
(Table~\ref{tab:littable})---the gap our own scope paragraph flagged as the
most consequential open question. We close it here: full fine-tuning of the
same student (fp32 weights, standard fp32 AdamW, batch and step budget
unchanged), on a grid $\{10^{-5}, 2{\cdot}10^{-6}, 10^{-6}\}$ that brackets
the exact published rates ($2{\cdot}10^{-6}$ in \citealp{jiang2026rock};
$10^{-6}$ in \citealp{xu2026tip}), $\{\textsc{Full}, \textsc{TV}\} \times 3$
seeds, verdict rule frozen in advance (Appendix~\ref{app:prereg}, P8). Two
engineering notes matter for validity: at these rates a single update moves
weights by less than bf16's resolution, so fp32 master weights are mandatory
and probed parameter displacements are asserted nonzero at run time
(Appendix~\ref{app:details}); and these runs live on different hardware with
its own re-measured baseline and a passed reproduction gate
(Appendix~\ref{app:prereg}, P8-anchor).

\begin{figure}[t]
\centering
\includegraphics[width=.72\linewidth]{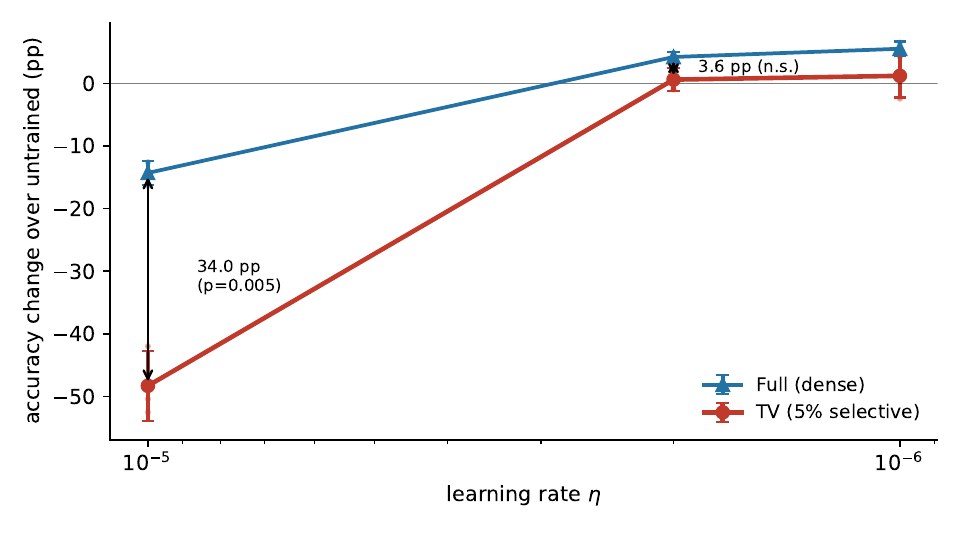}
\caption{\textbf{Under full fine-tuning the asymmetry survives in graded
form and grows.} Accuracy change over the untrained student (A800 baseline,
mean $\pm$ seed SD, $n{=}3$) at rates bracketing those the literature
publishes with. Dense swings $19.8\pp$; the TV-selective arm swings
$49.5\pp$ ($2.5\times$ more). The dense-versus-selective verdict is a
non-significant $+3.6\pp$ at the published operating point
($2{\cdot}10^{-6}$) and $+34.0\pp$ ($p{=}0.005$) one notch hotter.}
\label{fig:ft}
\end{figure}

\begin{table}[t]
\centering
\caption{\textbf{Full fine-tuning grid.} Accuracy change (pp) over the
untrained student on the same hardware ($63.46\%$; see
Appendix~\ref{app:prereg} for why this baseline differs from the LoRA
tables'), mean $\pm$ seed SD, $n{=}3$.}
\label{tab:ft}
\begin{tabular}{lrrrr}
\toprule
Arm & $\eta{=}10^{-5}$ & $2{\cdot}10^{-6}$ & $10^{-6}$ & Swing \\
\midrule
\textsc{Full} & $-14.30{\pm}1.9$ & $+4.22{\pm}0.9$ & $+5.53{\pm}1.2$ & $19.84$ ($p{<}0.001$) \\
\textsc{TV}   & $-48.32{\pm}5.6$ & $+0.61{\pm}1.9$ & $+1.21{\pm}3.5$ & $49.53$ ($p{=}0.001$) \\
\midrule
gap & $+34.02{\pm}3.4$ ($p{=}0.005$) & $+3.61{\pm}1.2$ ($p{=}0.060$) & $+4.32{\pm}2.1$ ($p{=}0.153$) & \\
\bottomrule
\end{tabular}
\end{table}

\paragraph{The entanglement reproduces and is $7\times$ larger.} The TV
arm's swing is $49.53\pp$ ($p{=}0.001$) against its LoRA counterpart's
$6.7$; at $\eta{=}10^{-5}$ full fine-tuning with a $5\%$ TV selection
destroys the model ($-48\pp$, final accuracy ${\sim}15\%$) while dense
training merely suffers ($-14\pp$). By the preregistered rule---TV swing
significant and larger than \textsc{Full}'s---the verdict is
\textsc{reproduced}.

\paragraph{An honest revision: dense is not flat here.} Under LoRA, dense
supervision was statistically flat and we leaned on that flatness. Under
full fine-tuning it is not ($19.84\pp$, $p{<}0.001$): at $10^{-5}$
\emph{everything} degrades. What survives, and what we now state as the
regime-independent form of the claim, is \emph{graded} asymmetry: in every
regime we measured, selective arms respond to the learning rate more than
dense ones---$2.5\times$ more here, unboundedly more under LoRA where the
dense response was indistinguishable from zero.

\paragraph{Published comparisons sit one notch from a cliff.} At the rates
this literature actually uses, our data shows selective TV training as
roughly neutral ($+0.6$ to $+1.2\pp$) and statistically indistinguishable
from dense at $n{=}3$ ($+3.6\pp$, $p{=}0.06$; $+4.3\pp$, $p{=}0.15$)---%
though at these cells' variance, $n{=}3$ has $80\%$ power only for effects
of ${\approx}4.4$ and $7.9\pp$ respectively, so these are bounds, not
equivalence claims. At
$5\times$ that rate, the same comparison reads $+34.0\pp$ ($p{=}0.005$).
Both facts matter: the published operating points are not obviously
misleading about selective methods' quality---but the verdict a shared-rate
comparison returns spans an order of magnitude within a $10\times$ rate
window, and nothing in the published protocols would detect this, because
none of them vary the rate. This is the strongest form of the paper's
thesis, obtained in the literature's own regime.

%======================================================================
\section{Implications for Selector Comparisons}
\label{sec:implications}

The prescription follows directly from the diagnosis, and it is two lines
long. First, tune the learning rate per arm and report the arm~$\times$~rate
matrix, not a shared-rate column: a selector comparison that fixes the
learning rate does not control for it, because every selective arm we
measured responds to the rate while dense training does not. Second, report the selection
drift $J_t$ alongside accuracy. It costs one extra forward pass of the
initial model per round, it is the observable on which the confound acts,
and a $J_t$ trajectory that differs across arms flags that selection is
responding to training. We offer it as a descriptive diagnostic: its link to
rate sensitivity is correlational in our data ($J_1$ orders with swing
across criteria), not a validated predictor.

\paragraph{Would a shared-rate selector comparison have been misled?}
The thesis concerns selector-versus-selector comparisons, so we run one, both
ways, on our own arms. Across the four selective criteria on the LoRA grid
the \emph{ranking} is stable---Entropy $>$ Random $>$ TV $>$ Teach at both
$10^{-4}$ and $5{\cdot}10^{-5}$---so in this setting the entanglement
stretches margins rather than reordering methods, and we report that as a
bounded negative. The stretching, however, is systematic and
conclusion-relevant: all six pairwise margins are $1.5$--$3.1\times$ larger
at the hotter rate, and two of the six flip significance. Entropy beats
Random by $+2.10 \pm 0.42\pp$ ($p{=}0.035$) at $5{\cdot}10^{-5}$ but by a
non-significant $+4.02 \pm 2.01$ ($p{=}0.16$) at $10^{-4}$; Random--TV
behaves the same way. A paper run at one shared rate concludes ``the
selector does not significantly matter''; the same paper at the other rate
concludes it does. No rank inversion is needed for the protocol to change
what a paper reports---the significance calls move first. (Scope: four
criteria, one dataset, $n{=}3$ per cell except TV's $n{=}12$.)

\paragraph{How the literature actually sets the rate.}
We checked the five selective-distillation papers closest to ours
(Table~\ref{tab:littable}). The pattern is uniform: where a learning rate is
reported at all, one value is shared by every selector arm, varied only by
model pair or training stage---never by method. \citet{xu2026tip} index
their hyperparameter table by model pair; \citet{jiang2026rock} give one
rate per training stage and state that no other hyperparameters were tuned;
\citet{koo2025switch} search a grid but by model size, and do not report
which value was selected. Two papers report no usable rate for their LLM
experiments at all. \emph{Nobody tunes the rate per selector arm}, which is
precisely the practice our results argue is unsafe.

\begin{table}[t]
\centering
\small
\caption{\textbf{Learning-rate practice in selective on-policy distillation.}
Rates as reported in each paper; ``shared'' means one value across all
selector arms compared in that paper.}
\label{tab:littable}
\begin{tabular}{llll}
\toprule
Paper & Reported rate & Per-arm tuning? & Adaptation \\
\midrule
\citet{xu2026tip}          & $10^{-6}$ / $3{\cdot}10^{-7}$ (by model pair) & no (shared) & full FT \\
\citet{jiang2026rock}      & $2{\cdot}10^{-5}$ off-policy, $2{\cdot}10^{-6}$ on-policy & no (shared) & full FT \\
\citet{koo2025switch}      & grid $\{5,1,0.5\}{\cdot}10^{-5}$ (by model size) & no (by size) & LoRA $r{=}16$ \\
\citet{huang2025selectkd}  & not reported for LLMs & --- & LoRA \\
\citet{wang2026teachability} & not reported & --- & not reported \\
\midrule
This work & grid $\{10,5,2.5,1.25\}{\cdot}10^{-5}$ & \emph{yes} (the point) & LoRA $r{=}32$ \\
\bottomrule
\end{tabular}
\end{table}

\paragraph{Is our grid in the right place?}
Two of these rates are one to two orders of magnitude below ours, and we
state plainly why the numbers are not directly comparable: those runs are
full fine-tunes, ours is LoRA, and LoRA is conventionally trained at
$10$--$100\times$ the full-fine-tuning rate. The like-for-like anchor is
\citet{koo2025switch}, the one LoRA-based comparison that reports a range:
its grid tops out at $5{\cdot}10^{-5}$, interior to ours. Our lower two grid
points thus sit inside published LoRA practice, and our upper point is one
notch hotter. And the full-fine-tuning regime is no longer an open question:
Section~\ref{sec:ft} runs the decisive contrast at the published rates
themselves, where the entanglement is $7\times$ larger than under LoRA and
the dense-versus-selective verdict moves from statistically invisible to
$34\pp$ within a $10\times$ rate window.

Read through this lens, the existing evidence base needs re-weighting rather
than dismissal. Our results do not say these selectors are without value;
they say a margin reported at one rate is one sample from a range that, in
our setting, spans a factor of $2.0$. The claims that survive this
re-weighting are the ones established across multiple rates; we found none
in this literature that report more than one.

\paragraph{Loop frequency is a second dial.}
An unplanned observation supports the same conclusion from another
direction. Two generations of our own pipeline differ in how often selection
is recomputed---every optimizer batch versus once per 512-step round---with
criterion, budget, rate, and steps held equal. At $\eta{=}5{\cdot}10^{-5}$
the per-batch variant lands at $+2.12\pp$ and the per-round variant at
$-2.15\pp$: a ${\sim}4.3\pp$ swing from the re-measurement frequency alone.
Seeds are unpaired across pipeline versions, so we report this as an
observation, not a claim; but it is exactly what a control-loop reading
predicts (re-measurement frequency changes loop behavior) and hard to
explain under the static-filter reading, where rescoring unchanged data
should be nearly idempotent.

\paragraph{Scope.}
Our evidence is two datasets (GSM8K throughout; MATH-500 for the decisive
contrast, where the entanglement did \emph{not} reproduce), one model pair
(\qwen{} 1.5B student,
7B teacher---a same-family configuration that \citet{li2026rethinkingopd}
identify as failure-prone, which is arguably the regime where selector
choice matters most), and two adaptation regimes (LoRA throughout; full fine-tuning for the
decisive dense-versus-selective contrast at the literature's published
rates, Section~\ref{sec:ft}). Three selector
criteria are covered (TV, entropy, teachability; Sections~\ref{sec:openloop}
and~\ref{sec:s2}), with the live/frozen decomposition available for all
three. The protocol-level entanglement claim holds for every selective arm
we tested; the additional contribution of the live scoring path is
established for TV at $n{=}12$ (Appendix~\ref{app:prereg}, P7) and, by the
S2 replication, does not extend to an entropy criterion.

%======================================================================
\section{Related Work}
\label{sec:related}

\paragraph{Hyperparameter tuning as a fairness control.}
That method comparisons can be decided by tuning effort rather than method
quality is an established concern outside distillation:
\citet{sivaprasad2020optimizer} show that optimizer rankings depend on the
hyperparameter-tuning protocol, and argue benchmarks must specify it. Our
contribution is to demonstrate the same failure inside selective
distillation, to locate it in an asymmetry between dense and selective
training rather than in tuning effort per se, and to give the diagnostic
(the arm~$\times$~rate matrix) that makes it visible.

\paragraph{Selective and token-level on-policy distillation.}
A growing line selects which positions of an on-policy rollout deserve
teacher supervision: by teachability \citep{wang2026teachability}, token
importance \citep{xu2026tip}, difficulty profile \citep{jiang2026rock},
weighted token subsets \citep{huang2025selectkd}, or scheduled teacher
involvement \citep{koo2025switch}; multi-teacher and dual formulations
extend the recipe \citep{ma2026mopd,yu2026dopd}. All of these recompute
selection from the live student---they are closed-loop in our sense---and
all evaluate selectors at a shared learning rate (Table~\ref{tab:littable}).
None vary the rate per arm or measure selection drift. Our contribution is orthogonal to each specific
criterion: it concerns the comparison protocol they share.

\paragraph{Phenomenology of on-policy distillation.}
\citet{li2026rethinkingopd} map when on-policy distillation fails, flagging
same-family small-gap pairs as a failure configuration, and
\citet{ma2026outcomeconfounded} analyze confounded supervision signals
within rollouts, and \citet{ma2026sekd} revisit what selective knowledge
distillation actually buys. Our finding that a \emph{random} $5\%$ subset is
already rate-entangled, and that all our selective arms lose to dense
supervision, is consistent with that sceptical line; what we add is that the
size of the loss depends on the rate, so the comparison protocol has to be
part of the discussion. We add a confound at the level of the experimental
protocol rather than the training signal, and our arm~$\times$~rate matrix
offers a re-reading of failure reports at a single rate: some ``failures''
may be loop instability at the chosen rate rather than properties of the
method.

\paragraph{Training systems that measure what they move.}
Feedback pathologies of this shape are documented elsewhere: recursive
training on model-generated data collapses the data distribution
\citep{shumailov2024collapse,alemohammad2024mad}, and optimizing against a
learned reward degrades as the policy consumes the signal that evaluates it
\citep{gao2023overoptimization}. Selective distillation has not previously
been placed in this family, and we stop short of claiming it belongs there:
our frozen-scoring ablation isolates the direct scoring path only, and its
effect on end-task accuracy is not established at our sample sizes.

%======================================================================
\section{Limitations and Conclusion}
\label{sec:conclusion}

\paragraph{Limitations.}
The strongest limitation is breadth: one model pair, and an entanglement
result that held on GSM8K but not on MATH---so the confound is demonstrated
to exist and to be large where it occurs, not to be universal. Whether the
difference is task-driven or power-driven (the MATH grid, at 325 evaluation
problems, could not have detected a $2$--$3\pp$ rate effect) is unresolved,
and a larger MATH evaluation set is the direct test. The
replication bounded one of our own claims---whatever the frozen-scoring
ablation detects is TV-specific---and the coupling account of
Section~\ref{sec:s2}, though it survived one frozen out-of-sample test, has
been tested on exactly one criterion with a wide interval. The
interaction between scoring source and rate is established but not precisely
located: $3.79 \pm 1.69\pp$ with a confidence interval whose lower end is
$0.29$, so ``live scoring roughly doubles the rate effect'' is the strongest
honest reading and a tighter estimate would need more seeds than the 48 we
ran. The load-bearing evidence in this paper is the
arm~$\times$~rate matrix and the within-run drift dynamics, and we have kept the two
categories separate throughout. The sign-flip result is a directional
observation, permanently capped at $n{=}5$ by our own stopping rule. The full fine-tuning replication
(Section~\ref{sec:ft}) removes the LoRA-only caveat for the central
contrast, though its frozen-scoring and drift instrumentation remain
LoRA-only. Finally, one sampling decision
was made after seeing a near-threshold statistic; its rules were frozen
before the new data, and both sample sizes appear in every affected table.

\paragraph{Conclusion.}
Selective distillation belongs to a family that machine learning keeps
rediscovering the hard way: training procedures that measure the thing they
move. The selector is not a lens held up to the data; it is a component
inside the loop whose stability the learning rate governs. The remedy is not
a better selector---it is a protocol that treats the loop as part of the
method: rates tuned per arm, drift reported alongside accuracy, and margins
read as what they are, upper bounds. A selector comparison that fixes the
learning rate does not control for it.

\bibliographystyle{plainnat}
\bibliography{refs}

\appendix
\section{Preregistration and Deviations}
\label{app:prereg}

This appendix is the audit trail: every decision criterion, frozen before
its data, and every deviation, with its cause.

\paragraph{P1 (Random falsification test)---fired.} Frozen before the
\textsc{Random} rate-grid runs: if the fully state-independent \textsc{Random}
arm swings more than $4\pp$ across the rate grid, the hypothesis that the
feedback loop is the \emph{sole} source of rate sensitivity is falsified; a
swing below $2.5\pp$ is consistent. On the original grid the measured swing
was $1.47\pp$ and the hypothesis survived---but a subsequent config audit
found that grid mixed training durations across columns (64 vs.\ 512 steps
per round), voiding the measurement. On the duration-homogeneous grid, with
both endpoint cells extended to $n{=}3$ under rules frozen before the new
data (thresholds unchanged; no further sampling), the swing is $5.08\pp$:
\emph{the criterion fired, and the strong hypothesis is recorded as
falsified}. The surviving two-layer account (support restriction entangles;
the loop deepens it) is what Sections~3 and~5 report.

\paragraph{P2 (frozen-scoring control, primary/secondary split).} Frozen before
the \textsc{TV-frozen} runs: drift trajectories are the primary observable
(within-run, ${\sim}512$ chains per round); the accuracy contrast was
declared underpowered in advance (predicted SE ${\approx}2\pp$) and demoted
to directional support regardless of outcome. The prediction: live-scored
$J_t$ dynamics depend on $\eta$, frozen-scored dynamics do not.

\paragraph{P3 (sample extension).} After the live-vs-frozen contrast at
$\eta{=}10^{-4}$ landed at $t{=}{-}2.77$ against a critical value of $2.78$
at $n{=}3$, we decided to extend---an optional-stopping situation, disclosed
as such. Rules frozen before the new data: (i) symmetric extension of both
closed cells to $n{=}5$, never only the near-threshold cell; (ii) the
identical Welch test, with critical values at the new degrees of freedom;
(iii) both $n{=}3$ and $n{=}5$ reported everywhere; (iv) no sampling beyond
$n{=}5$ regardless of outcome. Outcome: the near-threshold effect shrank
($-5.43 \to -2.40$), and the verdict ``directional support'' is permanent.

\paragraph{P4 (S2 prediction)---fired.} Frozen before the S2 runs, as
executable assertions in a hash-stamped verdict script: a student-only
entropy selector reads $\theta_t$ too; if the scoring-dependence hypothesis
holds as stated, its live arm must entangle more than its frozen arm (E1), and the drift-recovery signature must reproduce (E2). Outcome:
E1 \textsc{revise-loop} (live $3.16$, $p{=}0.043$; frozen $4.30$,
$p{=}0.059$; interaction $-1.14 \pm 1.34$, absent), E2 \textsc{not
reproduced} ($1/3$ seeds). The loop hypothesis as stated was wrong; the
criterion-scoped version and a post-hoc coupling account appear in
Section~\ref{sec:s2}. T1 (teachability entangles): $11.68 \pm 1.93$,
$p{=}0.016$, confirmed. All S2 cells are capped at $n{=}3$; no top-ups were
made.

\paragraph{P5 (coupling account, out-of-sample test)---passed.} After the P4
verdict we stated the churn-carrier account and froze its prediction for the
one unrun cell, in a hash-stamped script, before the data: the Teach
open/closed interaction must be indistinguishable from zero
($|I| \leq 2\,\mathrm{SE}$) and below TV's $+3.60$; an interaction
$\geq 3.60$ with $t{>}2$ refutes the account. Measured: $-6.07 \pm 5.06$
($t{=}{-}1.20$)---criterion met, account survives; the interval is wide and
we claim survival, not confirmation. One process note: our first informal
statement of this prediction had the sign of the ordering wrong
(``Teach's interaction should be largest''); writing the frozen criterion
caught the error before any data were collected.

\paragraph{P6 (dataset replication)---not reproduced.} Rule fixed before the
MATH runs: entanglement counts as reproduced only if \textsc{TV-live}
shows a significant rate effect that also exceeds \textsc{Full}'s; if
\textsc{TV-live}'s effect is non-significant \emph{and} no larger than
\textsc{Full}'s, the claim is scoped to GSM8K. Outcome: $1.74\pp$
($p{=}0.48$) versus $2.56\pp$ ($p{=}0.33$) $\Rightarrow$ scoped
(Section~\ref{sec:m500}). Unlike P4 and P5, this rule was recorded in the
project log rather than hash-stamped in a script before execution; we note
the weaker provenance rather than claim otherwise.

\paragraph{P7 (powered interaction test)---running.} An earlier version of
this paper argued that freezing the scoring model removes rate entanglement,
on the grounds that the rate effect was significant in the live arm
($p{=}0.032$) and not in the frozen arm ($p{=}0.25$). That is the
difference-of-significance fallacy \citep{gelman2006difference}; the
interaction it stands in for is $3.60 \pm 2.84\pp$ ($t{=}1.27$, $p{=}0.31$)
and establishes nothing. The claim was withdrawn and the test it needed was
preregistered in a hash-stamped script, frozen before any of its data
existed: all four cells of $\{\text{live}, \text{frozen}\} \times
\{10^{-4}, 5{\cdot}10^{-5}\}$ extended to $n{=}12$ (the size at which an
effect of the observed magnitude would reach $t{\approx}2$), Welch
interaction test, $p{<}0.05$ with positive sign confirms, anything else
demotes the effect permanently to the drift level, no top-ups regardless of
outcome. \textbf{Outcome: confirmed}---$3.79 \pm 1.69\pp$, $t{=}2.24$,
$p{=}0.035$, $95\%$ CI $[+0.29, +7.30]$, on 32 additional training runs.
The point estimate moved from $3.60$ to $3.79$ as $n$ went from $5/3$ to
$12$; it is the stability of that estimate, more than the $p$-value, that we
regard as the evidence. The claim is restored to the paper in the scoped
form Section~\ref{sec:openloop} states, and the invalid inference that
preceded it is left on the record here.

\paragraph{Statistical accounting.} The preregistered decision criteria
above (P1--P6) are the confirmatory tests; every other $t$-statistic in the
paper---the ladder contrasts of Section~\ref{sec:entangle}, the post-hoc
interaction term, the per-cell comparisons---is descriptive and reported
without multiplicity correction. We state this rather than apply a
correction after the fact, since the confirmatory set was fixed in advance
and is small.

\paragraph{P8 (full fine-tuning replication)---reproduced.} Rules frozen in
the project log before any full-fine-tuning data existed, verdict script
written before the final 2 of 18 runs completed: grid $\{\textsc{Full},
\textsc{TV}\} \times \{10^{-5}, 2{\cdot}10^{-6}, 10^{-6}\} \times 3$
seeds; \textsc{reproduced} iff TV's swing is significant (Welch $p{<}0.05$)
\emph{and} exceeds \textsc{Full}'s; $n{=}3$ final, no top-ups. Outcome: TV
$49.53\pp$ ($p{=}0.001$) vs.\ \textsc{Full} $19.84\pp$
($p{<}0.001$)---\textsc{reproduced} (Section~\ref{sec:ft}).

\paragraph{P8-anchor (hardware reproduction gate)---passed.} The full
fine-tuning runs live on different hardware (A800 80\,GB; all earlier runs:
RTX 4090). Gate, frozen in advance: the LoRA TV arm re-run on the new
machine must swing $\geq 4\pp$ across $\{10^{-4}, 5{\cdot}10^{-5}\}$,
or rate sensitivity is machine-dependent and the full-FT results are
uninterpretable. Measured: swing $6.04$ ($-8.19 \pm 2.0$ vs.\ $-2.15 \pm
1.5$, $n{=}3$), against $6.68$ ($-8.18$, $-1.50$) on the original
machine---the training effect reproduces across GPUs almost exactly.

\paragraph{Hardware sensitivity of the evaluation itself.} The new machine
required its own baseline: the identical checkpoint, prompt (SHA-verified),
generation config (SHA-verified), dataset (hash-verified), torch and
transformers versions scores $63.46\%$ on the A800 against $64.97\%$ on
the RTX 4090---a $1.51\pp$ difference from the GPU alone. We traced it to
single-ULP bf16 logit differences at near-tied top-2 positions, which greedy
decoding then amplifies autoregressively; \citet{yuan2025nondeterminism}
document the same mechanism at up to $9\%$ across GPU types. Two
consequences for this paper: all full-FT numbers are reported against the
A800 baseline and never numerically differenced against 4090 results; and
the observation itself reinforces the thesis, since a hardware swap moves
the measured score by more than a typical published selector margin.

\paragraph{Deviations and recorded errors.} Three times during this project
a criterion written in prose was translated into analysis code incorrectly
(a drift sanity check whose ``passing'' value was actually the failure
signature; a three-branch verdict encoded as two branches; a threshold
mis-transcribed). All three were caught by cross-checking code against the
frozen prose. The resulting practice, adopted midway: criteria are frozen
\emph{as executable assertions}, not as prose to be translated later. We
report this because protocols fail at the translation step more often than
at the design step.

\paragraph{Data hygiene.} Two result files from early runs were overwritten
by later runs before an output-tagging flag existed; both were detected by
config-fingerprint audits of every artifact (the values survive in logs) and
neither enters any table in this paper. Runs that mixed training durations
(64 vs.\ 512 steps per round) across cells of the rate matrix were likewise
detected by audit and excluded. Every number in this paper comes from the 133
duration-homogeneous runs (1{,}536 steps each) whose configs are fingerprinted
in the released artifacts. One further bug is disclosed because it shaped an
intermediate analysis: the training script computes ``change over untrained''
against a single cached baseline file, which is dataset-specific; the MATH
runs therefore initially reported deltas against the GSM8K baseline. All
MATH numbers in this paper are raw accuracies, which the bug does not touch,
and all MATH contrasts are within-dataset differences, in which the constant
cancels.

\section{Training and Measurement Details}
\label{app:details}

\paragraph{Training.} LoRA rank 32 on all attention and MLP projections;
AdamW; gradient clipping at $1.0$; batch size 2; 512 rollouts per round from
GSM8K training prompts; 3 refresh rounds $\times$ 512 optimizer steps;
distillation loss on teacher top-512 support (computed as $\mathrm{logit} -
\mathrm{logsumexp}$ without materializing full-vocabulary log-softmax).
Generation: 768 max new tokens, temperature $0.7$, top-$p$ $0.9$, matching
the configuration under which all diagnostic surfaces were measured. The
teacher is released from memory after scoring each round.

\paragraph{Evaluation.} Greedy decoding on all 1{,}319 GSM8K test problems,
final-number extraction. Sampled evaluation differs from greedy by up to
$25\pp$ at this model scale and is unusable for contrasts of a few pp.

\paragraph{Update-size measurement.} For every optimizer step we log the
pre-clip gradient norm and the actual parameter displacement
$\lVert\theta_{k+1}-\theta_k\rVert$. The $2.2\%$ figure in
Section~\ref{sec:notstepsize} is the across-arm dispersion of
displacement$/\eta$. Master weights are fp32; bf16 master weights silently
swallow displacements three orders of magnitude below their resolution, a
failure mode we hit and instrumented against (an assertion verifies that
applied steps change the weights).

\paragraph{Scoring-source assertions.} The open/closed switch is validated
structurally, not by the drift statistic: the training mask must equal the
mask induced by the declared scoring source, and once the model has moved,
must differ from the mask induced by the other source. Both checks are hard
failures at run time. (The tempting alternative---checking that frozen-scored
drift stays at $1.0$---is backwards: drift compares $\theta_t$ to $\theta_0$
scores regardless of which one selects, and a drift pinned at $1.0$ is
precisely the signature of a switch that silently does nothing.)

\paragraph{Full fine-tuning specifics.} fp32 master weights are mandatory,
not preferential: at $\eta = 10^{-6}$ a single AdamW step moves a weight by
${\sim}10^{-6}$ against magnitudes of ${\sim}10^{-2}$---a relative change
of $10^{-4}$, two orders of magnitude below bf16's ${\sim}4{\cdot}10^{-3}$
resolution, so under bf16 every update would be silently rounded away and
the run would produce a clean-looking flat result. Guards: parameter
displacement is probed on a fixed 8-tensor subset every 32 steps (cloning
all $1.5$B parameters per step, as the LoRA path does with its $37$M, would
cost $6.2$\,GB per step), any probed step with zero movement is a hard
failure, and a run ending with no observed movement aborts rather than
reporting. Optimizer states are fp32 AdamW (no quantization); peak memory
$32.2$\,GB on an 80\,GB A800. The frozen-scoring arm is unavailable under
full fine-tuning ($\theta_0$ is not recoverable without a resident copy)
and the drift statistic is recorded as undefined rather than approximated.

\paragraph{Compute.} All runs on a single RTX 4090 (24 GB); a full
three-round run takes ${\sim}75$ minutes. The complete evidence base of this
paper is ${\sim}205$ GPU-hours across two machines, including audits and
discarded grids.

\end{document}